\pdfoutput=1

\documentclass[11pt]{article}

\usepackage{emnlp2021}

\usepackage{times}
\usepackage{latexsym}

\usepackage[T1]{fontenc}

\usepackage[utf8]{inputenc}

\usepackage{microtype}

\usepackage{graphicx}
\usepackage{amsmath}
\usepackage{amsfonts}
\usepackage{xcolor}
\usepackage{enumitem}
\usepackage{multirow}
\usepackage{booktabs}
\usepackage{color, colortbl}
\definecolor{GreenRow}{RGB}{172,204,186}

\title{Mapping Language to Programs using Multiple Reward Components with Inverse Reinforcement Learning}

\author{Sayan Ghosh \;\;\;\;\;\; Shashank Srivastava\\ 
  UNC Chapel Hill\\ 
  \texttt{\{sayghosh,ssrivastava\}@cs.unc.edu}
}

\begin{document}
\maketitle

\begin{abstract}
Mapping natural language instructions to programs that computers can process is a fundamental challenge. Existing approaches focus on likelihood-based training or using reinforcement learning to fine-tune models based on a single reward. In this paper, we pose program generation from language as Inverse Reinforcement Learning. We introduce several interpretable reward components and jointly learn (1) a reward function that linearly combines them, and (2) a policy for program generation. Fine-tuning with our approach achieves significantly better performance than competitive methods using Reinforcement Learning (RL). On the VirtualHome framework, we get improvements of up to 9.0\% on the Longest Common Subsequence metric and 14.7\% on recall-based metrics over previous work on this framework~\cite{puig2018virtualhome}. The approach is data-efficient, showing larger gains in performance in the low-data regime. Generated programs are also preferred by human evaluators over an RL-based approach, and rated higher on relevance, completeness, and human-likeness.
\end{abstract}

\section{Introduction}
\begin{figure}[!t]
    \centering
    \includegraphics[scale=0.28]{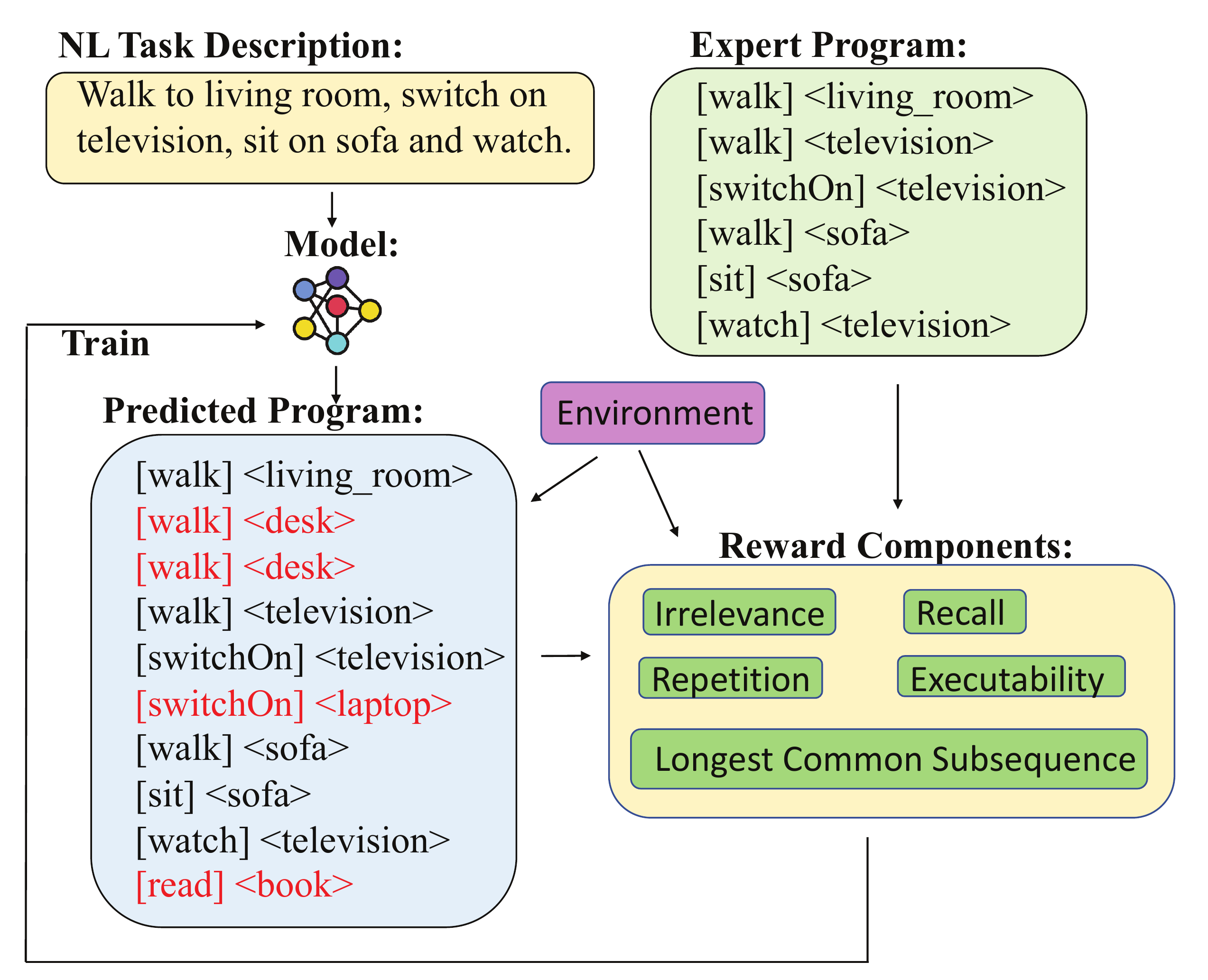}
    \caption{We frame conversion of a natural language task description into a program that can be executed in an environment as Inverse Reinforcement Learning. We design multiple interpretable reward components encoding preferred characteristics in generated programs. A reward function (optimally combining these components) and a policy for program generation are jointly learnt from expert demonstrations/programs. 
    }
    \label{fig:task_figure}
\end{figure}
Mapping natural language descriptions to programs is vital for developing agents that can mimic human behavior in the real world. For example, imagine a robot that is instructed to watch television (Figure \ref{fig:task_figure}). The robot needs to parse the language instruction into a sequence of commands for navigating to the living room, then going to the sofa, turning on television, etc. For parsing, the robot needs to map objects mentioned in the instruction to objects in the surrounding environment, and then perform the actions required to complete the task. Parsing language into actions has been widely explored in diverse settings \cite{chen2011learning, chen2019touchdown, anderson2018vision, tellex2011understanding}. Most recent approaches leverage supervised learning with maximum likelihood estimation, followed by a fine-tuning phase of reinforcement learning with a single user-specified reward, encoding signals for things like task completion and executability~\cite{misra2017mapping, goyal2019using}. To alleviate the problem of sparse rewards, reward shaping terms are often used to encode aspects like distance from goal state or deviation from labeled trajectories~\cite{misra2017mapping}. Reward signals and shaping terms are combined using hyperparameters which need to be manually tuned, and this process becomes tedious. Thus, these approaches can't efficiently leverage multiple possible sources of supervision that might be available. This paper explores an approach to alleviate this problem.  

Figure \ref{fig:task_figure} illustrates the core idea of this paper. We use multiple reward components encoding various desired characteristics of a good program to drive program generation, given a natural language task description and an environment. Each component consists of a manually defined reward (such as irrelevance, recall, and repetition) with clear semantics. The example in the figure illustrates the need for multiple reward components. We note that the predicted program mentions all the commands in the same sequence as in the expert program. However, it also generates irrelevant commands and repeats steps. Thus even though it scores optimally in the Longest Common Subsequence metric which has been used as a reward for this task \cite{puig2018virtualhome}, we need other signals to improve program generation. 
We investigate subsuming such reward components in an IRL framework to jointly learn: (1) a composite reward function combining the reward components with optimal weights, and (2) a policy that closely mimics the expert. The only supervision involved consists of task descriptions paired with 
labeled (expert) programs. Automatically learning the weights of reward components becomes increasingly vital as their number increases.

In terms of testbeds, much of previous work on instruction following has focused on block worlds \cite{misra2015environment} and navigation tasks \cite{chen2019touchdown, anderson2018vision, misra2018mapping, ALFRED20}.  We focus on the VirtualHome environment~\cite{puig2018virtualhome}. In contrast with previous datasets, it contains stepwise instructions and programs for a large number of realistic household activities, such as making coffee or folding laundry. Steps in the programs often involving interacting with objects, or changing the state of the environment. Thus, it offers a rich set of realistic challenges including object interactions and commonsense reasoning for program generation. 

Our evaluation shows that using IRL to fine-tune a model leads to significant gains over reinforcement learning. In fact, IRL leads to improved performance even on some metrics that a reinforcement learning policy directly optimizes. Additionally, our approach is data efficient. Experimenting with different  dataset sizes reveals that the method generalizes better than baseline methods in low-data scenarios. More significantly, the approach can extend to other domains and provides a general framework for incorporating multiple sources of supervision or inductive biases about a task.

Our contributions are:
\begin{itemize}[noitemsep, topsep=0pt, leftmargin=*]
    \item We pose mapping task descriptions to programs as an IRL problem, and learn a composite reward function combining semantically interpretable characteristics of expert programs\footnote{Code and dataset splits for the paper are available at \url{https://github.com/sgdgp/VirtualHome_IRL}}. 
    \item We achieve up to 9\% increase in the Longest Common Subsequence (LCS) metric w.r.t. previous methods. Programs generated by our approach are qualitatively better and are preferred by human evaluators. 
    \item Our approach is data-efficient in limited data scenarios compared to previous methods.
\end{itemize}

\section{Related Work}
\textbf{Semantic Parsing and Instruction Following:} 
Parsing natural language to programs has been explored in diverse settings. Common semantic parsing applications include text-to-SQL \cite{zhong2017seq2sql, spider, sparc, cosql} text-to-code \cite{yin2018mining, shin2019program}, robot navigation and interaction tasks \cite{misra2016tell, nyga2018grounding, squiregrounding}. Other tasks involve mapping instructions to actions in simple environments \cite{artzi2013weakly,chen2011learning,misra2015environment,malmaud2014cooking}. Instruction following is also a crucial part of complex Vision and Language Navigation (VLN) tasks \cite{anderson2018vision,misra2018mapping,chen2019touchdown,nguyen2019help}. Significant work has explored developing models that can use additional context in this space \cite{fried2018speaker,ke2019tactical,wang2019reinforced, nguyen2019help, nguyen2019vision,thomason2020vision,ma2019self, ma2019regretful}. Compared to environments like Room2Room~\cite{anderson2018vision}, VirtualHome~\cite{puig2018virtualhome} 
contains realistic activities and involves dynamic state changes and interacting with the objects to complete tasks. \\
\noindent \textbf{Training paradigms:} 
Encoder-decoder architectures~\cite{sutskever2014sequence} are the dominant modeling paradigm for instruction following tasks. 
Models are usually pre-trained with Maximum Likelihood Estimation, and fine-tuned with Reinforcement Learning \cite{puig2018virtualhome, misra2017mapping, zhong2017seq2sql} using REINFORCE~\cite{williams1992simple}. 
There has been limited work on using IRL~\cite{abbeel2004apprenticeship, ziebart2008maximum, finn2016guided, shi2018toward, li-etal-2018-paraphrase, sayanzhengacl} in NLP. \newcite{fu2018from} use IRL to learn a language-guided reward by using a deep neural network to parameterize a reward function.
Our work is closest to \newcite{sayanzhengacl}, which formulates reward components for table-to-text generation and learns their optimal linear combination using IRL. Following similar ideas, we define multiple interpretable reward components representing desired characteristics of good programs in our context and learn their optimal linear combination while jointly learning a policy for program generation. However, we differ from \newcite{sayanzhengacl} by applying IRL on a different downstream task (program generation from natural language instruction) and further show that using IRL as opposed to RL can lead to higher gains in limited labelled data scenarios.

\section{Method}
Each example in our data consists of a natural language description of a task to be performed in an environment, along with its corresponding program (see Figure \ref{fig:task_figure}). In VirtualHome, a program is a sequence of commands that can be executed to complete the task being described. A command is an action-object-object triplet, where one or both of the object arguments can be empty based on the action. For example, in the command ``\textit{[walk] $\langle$living\_room$\rangle$}" the action ``\textit{walk}" is followed by an object argument ``\textit{living\_room}". Similarly the command ``\textit{[putBack] $\langle$plate$\rangle$ $\langle$cupboard$\rangle$}" has two object arguments and ``\textit{sleep}" has none.

Let $D = \{d_i ; 1 \le i \le m\}$ 
denote the natural language description of the task to be performed in VirtualHome, where $d_i$ denotes the $i^{th}$ token. Let $C = \{ c_j ; 1 \le j \le n \}$ denote the corresponding program, where $c_j$ denotes $j^{th}$ command. Our objective is to predict program $C$ given the text description $D$ and an environment $E$. Previous approaches performed likelihood-based pre-training followed by fine-tuning using reinforcement learning \cite{puig2018virtualhome}. Instead, we use IRL to learn the reward function from the expert programs and fine-tune the program generation policy simultaneously. We design a set of rewards with clear semantics that encode the desired characteristics of a good program in our context. In the rest of this section, we describe the model architecture, provide definitions of reward components, and finally describe model training using IRL. 

\begin{figure*}[htbp]
\small
    \centering
    \includegraphics[scale=0.45]{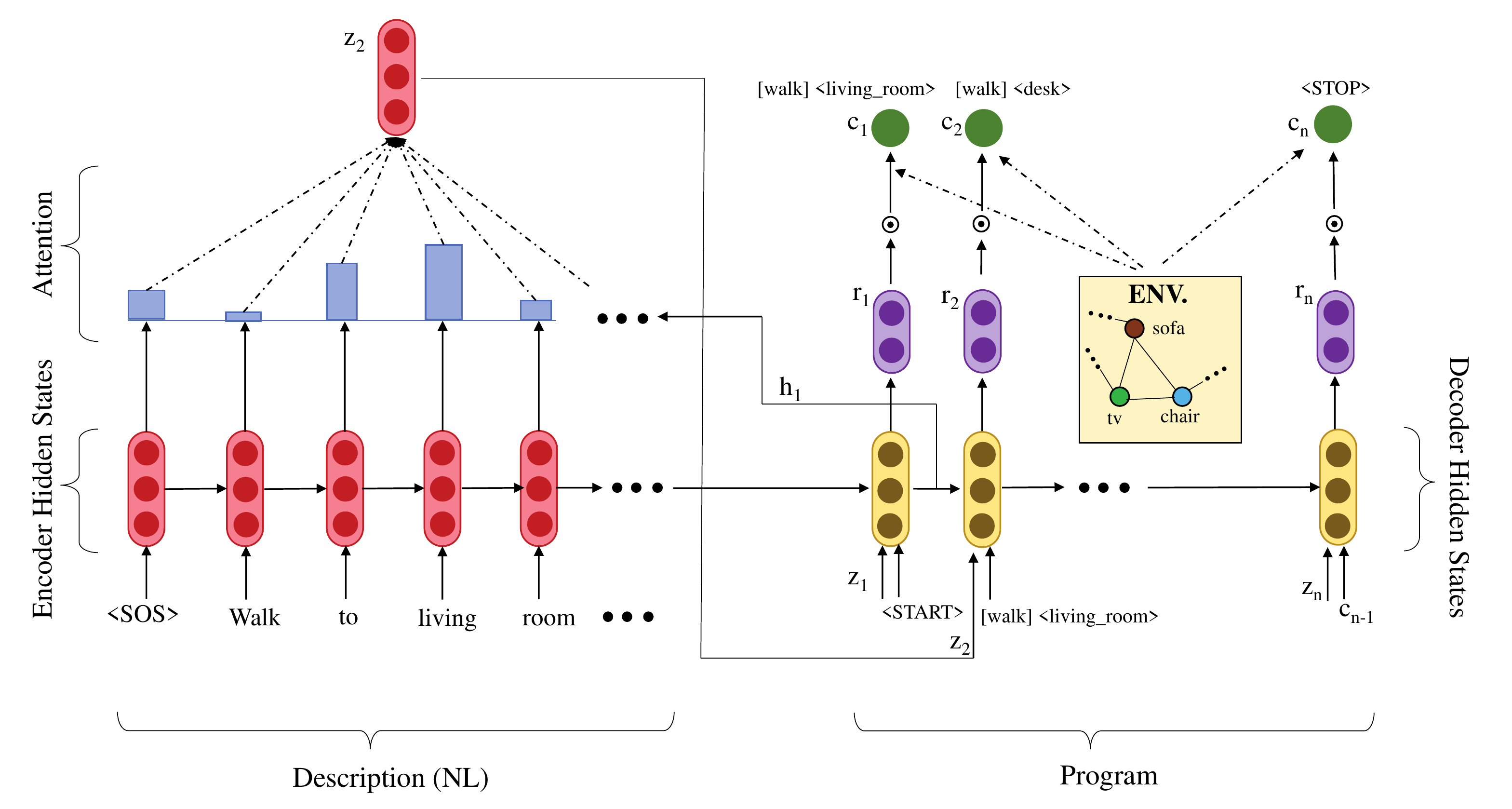}
    \caption{We follow a similar architecture to \citet{puig2018virtualhome}. The final hidden state of the encoder is used to initialise the hidden state of the decoder. Each decoder cell takes in the previous step command embedding($\psi(c_{t-1})$) and a context vector $z_{t}$(computed by attention over encoder hidden states) as input. The output of the decoder($r_{t}$)  is used to predict the next command by computing cosine similarity (denoted by $\odot$) to embeddings of all possible commands. During training, the most probable action is chosen. While testing, we explore two variations of inference : (1) regular beam search, and (2) beam search enforcing executability as a hard constraint. 
    }
    \label{fig:m1}
\end{figure*}

\subsection{Model architecture}
We use an encoder-decoder architecture (illustrated in Figure~\ref{fig:m1}). 
The encoder is modeled as a RNN with LSTM cells, and provides a representation of the task description. The decoder is also modeled by using LSTM cells for predicting commands at each time step. We pre-compute a representation/embedding for every possible command by simply averaging the word2vec \cite{word2vec} embeddings of its action and object arguments. Let $\psi$ denote the function that maps a command to its embedding. \\
\noindent{\textbf{Description encoder:}}
Task descriptions are tokenized, and each token represented using its word2vec embedding. 
Tokens are passed through a LSTM network to get the representation of the description. The final hidden state of the encoder is used to initialize the hidden state of the decoder. \\
\noindent{\textbf{Attention over text encoder:}}
We attend over the sequence of encoder hidden states, $h^{enc}$, for every step $t$, using the previous hidden state of the decoder,  $h_{t-1}$. We obtain the context vector, $z_t$, as:

\begin{gather}
    \alpha^t = \text{softmax} (\psi(c_{t-1})^T W_{att} [h_{t-1},h^{enc}])\\
    z_t = \sum_{j=1}^{m} \alpha^t_j h^{enc}_j 
\end{gather}
where $\psi(c_{t-1})$ is the embedding for the previous command and $ W_{att}$ is a learnable matrix.  \\

\noindent{\textbf{Decoder:}}
The decoder is an RNN with LSTM cells. 
The LSTM takes as input a concatenation of the previous command embedding, $\psi(c_{t-1})$, and a context vector, $z_t$. 
 The operation of the decoder at step $t$ can be described as :
\begin{equation}
    r_t = W_{dec} \times LSTM([\psi(c_{t-1}), z_{t}] , h_{t-1}, s_{t-1})
\end{equation}
where $h_{t-1}$ and $s_{t-1}$ are the hidden and cell states at step $t-1$ and $W_{dec}$ is a learnable weight matrix that defines a linear transformation from LSTM output space to command-embedding space. \\ 
\noindent{\textbf{Command prediction:}}
The decoder's output, $r_t$, is a vector in the command-embedding space. We calculate its cosine similarity with each possible command in the environment. The most similar command is chosen as the output at every step.

\subsection{Reward components}
\label{sec:reward_comp}

Defining rewards for reinforcement learning requires identifying preferred characteristics of outputs. 
Intuitively, correct programs corresponding to a task description have multiple characteristics: they should contain objects and actions similar to those mentioned in the task description in their arguments and actions, but not contain many unmentioned objects and actions. Moreover, the order of the commands should be semantically feasible. Also, the program should not repeat cycles of commands (a common problem in conditional generation models) and be executable in the given environment. Finally, these programs should be similar to expert programs. A key insight is that these ideas can be captured quantitatively with reward components, which we describe next. 
In the following, $C^{pred}$ and $C^{gt}$ denote the predicted and ground truth (expert) program. 
$O^{pred}$ denotes the set of objects in $C^{pred}$ and $O^{gt}$ denotes the set of objects (nouns) in the task description. 

\begin{itemize}[noitemsep,topsep=0pt,leftmargin=*]
    \item \textbf{Recall from description}: 
    An object in $O^{pred}$ is similar to some object in $O^{gt}$ if their cosine similarity is above a threshold. This reward value is the fraction of objects in $O^{gt}$ similar to any object in $O^{pred}$. 
    
    \item \textbf{Irrelevance}: 
    For a command predicted at step $t$ if the object present in the command is not relevant w.r.t. $O^{gt}$ 
    a penalty of -1 is given to this command. An object in the command is related to the description if its cosine similarity with at least one object from $O^{gt}$ is greater than a threshold. The total penalty for a predicted program is normalized by its length. 
    \item \textbf{Repetition}: We penalize the programs for repetition. For every predicted command $c_t$ (for $t >= 2$) if the bigram $(c_{t-1}, c_{t})$ is not unique, a penalty of -1
    is given to $c_t$. The total penalty is normalized by the length of the sequence. 
    \item \textbf{Longest Common Subsequence (LCS)}: We quantify how close the predicted program is to the expert program by finding the LCS score between $C^{pred}$ and $C^{gt}$ normalized by length. 
    
    \item \textbf{Recall from program}: 
    Fraction of the set of commands in $C^{gt}$ contained in $C^{pred}$.
    \item \textbf{Executability} : Program is given reward (+1) if it is executable in the given environment. 
    
\end{itemize}

The total reward for a program is a linear combination of the above components.

\subsection{Training}
We pre-train the model in a supervised manner, followed by finetuning it by using Maximum Entropy IRL. Details on implementation and hyper-parameters are provided in the supplementary material.
\subsubsection{Supervised training}
We optimize for the maximum likelihood estimation (MLE) objective using a cross-entropy loss at each step. We use teacher-forcing~\cite{bengio2015scheduled} during training, by using ground truth commands at a step as decoder inputs for the next step. 
\subsubsection{Maximum Entropy (MaxEnt) IRL }
We fine-tune the model using MaxEnt IRL~\cite{ziebart2008maximum} to mimic salient characteristics of expert demonstrations. We formulate program generation as IRL, where at each step $t$, a command is generated. The reward is not observed but learned from expert demonstrations in the training data. IRL consists of two alternating steps: (1) \textbf{Reward approximation:} estimate the underlying reward function using the expert demonstration and the current policy.
(2) \textbf{Reinforcement Learning:} use the estimated reward function to optimize the policy for program generation.

\noindent \textbf{Reward approximation:} Following the standard MaxEnt IRL framework, we assume that expert programs are drawn from a distribution  $p_\phi(C|D,E)$.
\begin{multline}
\small
    p_\phi(C|D,E) = \frac{1}{Z}\exp(R_\phi(C|D,E)) \\
    \text{ and } Z = \int_{C}\exp(R_{\phi}(C|D,E))
\end{multline}
where the reward function, $R_\phi(C|D,E)$ has parameters $\phi$,  and $Z$ is the partition function. Also, the total reward of a program is sum of rewards at each step. 
Given a policy for program generation $q_{\theta}(C|D,E)$, our objective is to maximise the log-likelihood of the samples in the training set (Equation \ref{eqn:obj}). 
\begin{multline}
\small
\label{eqn:obj}
    J_r(\phi) = \frac{1}{N}\sum_{n=1}^{N}\log(p_\phi(C_n|D_n,E_n)) \\ = \frac{1}{N}\sum_{n=1}^{N}R_\phi(C_n|D_n,E_n) - \log Z
\end{multline}
Thus, gradient w.r.t. reward parameters is given by 
\begin{multline}
\small
    \nabla_{\phi}J_r(\phi) = \frac{1}{N}\sum_{n}\nabla_\phi R_\phi(C_n|D_n,E_n)\\- \frac{1}{Z}\int_{C}\exp(R_\phi(C|D,E))\nabla_\phi R_\phi(C|D,E) dC \\
    = \mathbb{E}_{C \sim p_{data}}\nabla_\phi R_\phi(C|D,E) - \\ \mathbb{E}_{C \sim p_{\phi(C|D,E)}} \nabla_\phi R_\phi(C|D,E) 
\end{multline}

\noindent We use importance sampling to approximate the gradient of the log partition function when drawing programs from the distribution of generated programs. The importance weight $\beta_i$ for a generated program $C_i$ is given by
\begin{equation}
    \beta_i \propto \frac{\exp(R_{\phi}(C_i|D_i,E_i))}{q_{\theta}(C_i|D_i,E_i)}
\end{equation}
Using importance sampling, the gradient of the objective function can be approximated as 
\begin{multline}
\small
    \nabla_{\phi}J_r(\phi) = \frac{1}{N}\sum_{i=1}^{N}\nabla_{\phi}R_\phi(C_i|D_i,E_i) - \\ \frac{1}{\sum_j \beta_j}\sum_{j=1}^{M}\beta_j \nabla_{\phi}R_\phi(C'_j|D'_j,E'_j)
\end{multline}
where $C_i$ and $C'_j$ are drawn from training data and $q_{\theta}(C|D,E)$ respectively.
In this work, we also assume that $R_{\phi}$ 
is a linear combination of the reward components defined in Section \ref{sec:reward_comp}. 
\begin{equation}
\small
    R_{\phi}(C|D,E) = \sum_{t=1}^{\tau} \phi^T \Psi^t
\end{equation}
where $\phi$ is a weight vector, $\Psi^t$ is the vector of reward component values at step $t$ and $\tau$ denotes total time-steps.
Owing to the linear formulation, the weight update for each reward component simply becomes a difference between the expected expert and the expected roll-out reward component. We use N expert programs and sample M programs from our policy $q_{\theta}(C|D,E)$. The weight update for a component $\psi$ is:
\begin{equation}
\small
    \nabla_{\phi}J_r(\phi)_{\psi}  =  \frac{1}{N}\sum_{i=1}^{N}{\psi_i} - \frac{1}{\sum_j \beta_j}\sum_{j=1}^{M}\beta_j {\psi^{'}_{j}} 
\end{equation}
where $\psi_i$ and $\psi^{'}_j$ are total value of reward component over all steps for $i^{th}$ expert program $j^{th}$ generated program respectively.
These weights are learned in a data-driven approach when the supervised model is fine-tuned using MaxEnt IRL. 

\noindent \textbf{Reinforcement Learning:} For reinforcement learning stage of IRL to learn the policy for program generation we use 
Self Critical Sequence Training (SCST) \cite{scst}. 
We also
perform entropy regularisation \cite{williams1992simple, nachum2017bridging} when training our model. 
The objective of the program generator ($J_g(\theta)$) is 
\begin{multline}
    J_g(\theta) = \mathbb{E}_{C \sim q_\theta(C|D,E)} [R_{\phi}(C|D,E)] + \\ \delta H(q_{\theta}(C|D,E))
\end{multline}
where $\delta$ is a hyper-parameter (0.05 for our experiments)
and $H(q_{\theta}(C|D,E))$ is the entropy of $q_{\theta}(C|D,E)$ .

\noindent
\textbf{Summary of training process:} The model training consists of an iterative process with two steps. In the first step, we fix the program generation policy, and use programs sampled from the current policy to update the weights corresponding to each reward component. For this step, we assume that the expert programs' distribution is exponential in their rewards. The sampled programs from the current policy are used to approximate the log partition function required in making the updates to the reward components. In the second step, we fix the composite linear reward function (i.e. the weights of each reward component), and use it to update the program generation policy for the current reward function using policy-gradient updates.

\section{Experiments and Results}
In this section, we describe the dataset, metrics used for quantitative evaluation followed by a comparative evaluation of the proposed method with baselines and results of human evaluation\footnote{Code and data splits are available at \url{https://github.com/sgdgp/VirtualHome_IRL}}.
\subsection{Data}
 The VirtualHome dataset consists of programs for household tasks, accompanied by task descriptions and environments with necessary preconditions to carry out the tasks. 
 VirtualHome provides a graph-based simulator, where each node represents entities (like rooms, object etc.) and the simulator tracks changes of attributes and interactions. 
 However, there are some programs not executable in any home scenario. 
 We remove these, and also keep a single program when multiple programs appear against a single task description.
 Thus our modified dataset consists of unique executable programs which is divided into train, validation, and test splits of sizes 697, 185 and 500 samples respectively. To study the performance of our method in a low data regime, we randomly sample and form two smaller subsets of the training set of sizes 221 and 70.
 To calculate rewards, we extract nouns from task descriptions using spaCy's POS tagger. 
\subsection{Metrics}
We use the following for quantitative evaluation.
\begin{itemize}[noitemsep,topsep=0pt,leftmargin=*]
    \item \textbf{Normalized Longest Common Subsequence (LCS):} 
    This is the length of the longest common subsequence of the predicted and the expert programs normalized by the maximum of their lengths.
    We find LCS values of actions, objects, and commands
    separately. The mean LCS score is the average of these. 
    
    \item \textbf{Edit Distance (ED):}  This metric denotes edit distance between the predicted and the expert programs normalized by the maximum of the two programs' lengths. 
    
    
    \item \textbf{Recall from program :} We calculate the average reward value for recall from program reward component as described in Section \ref{sec:reward_comp}. 
    
    \item \textbf{Executability :} 
    Using the available preconditions for the programs, we obtain the percentage of generated programs which are executable in the graph-based simulator of VirtualHome.  
\end{itemize}
For all metrics except \textbf{ED}, a higher value is better.
However, note that our metrics do not explicitly capture task completion. For example, the executability metric will be 1 even if the task is incomplete as long as the predicted program is executable in the environment. 
\subsection{Quantitative Evaluation}
\begin{table*}[!ht]
\centering
\small
\begin{tabular}{lcccc}
\toprule
                                 \textbf{\textsc{Model}}   & 
                  \textbf{\shortstack{ \textsc{Mean}\\\textsc{LCS}}} &
                  \textbf{\textsc{ED}} &
                  \textbf{\shortstack{\textsc{Recall} \\ \textsc{Prog.}}} &
                  \textbf{\textsc{Exec.}} 
                  \\ 
\midrule
Random &  0.075 $\pm$0.002 & 0.997 $\pm$0.000 & 0.009 $\pm$0.004 & 0.003 $\pm$0.003\\    
\midrule
\rowcolor{GreenRow}
\multicolumn{5}{c}{\textbf{697 training samples}}\\


MLE & 0.368 $\pm$0.015 & 0.750 $\pm$0.016 & 0.357 $\pm$0.023 & 0.348 $\pm$0.045 \\
RL (LCS) & 0.400 $\pm$0.044 & 0.715 $\pm$0.046 & 0.404 $\pm$0.067 & 0.379 $\pm$0.055 \\ 
RL (LCS)$^\diamond$ & 0.410 $\pm$0.009 & 0.701 $\pm$0.014 &0.423 $\pm$0.015 & \textbf{1.000} $\pm$0.000 \\
RL (LCS + exec) & 0.385 $\pm$0.055 & 0.731 $\pm$0.056 & 0.373  $\pm$0.074 & 0.373 $\pm$0.081 \\

RL (LCS + exec)$^\diamond$ & 0.372 $\pm$0.065 & 0.737 $\pm$0.066 & 0.363 $\pm$0.088 & \textbf{1.000} $\pm$0.000 \\
RL (All - Exec.) &  
 0.431 $\pm$0.007 & 0.694 $\pm$0.013 & \textbf{0.460} $\pm$0.015 & 0.392 $\pm$0.025 \\

RL (All) & 0.423 $\pm$0.016 & 0.698 $\pm$0.017 & 0.448 $\pm$0.022 & 0.394 $\pm$0.027 \\

Ours (IRL: All - Exec.) & \textbf{0.436$^{*}$} $\pm$0.012 & 0.687 $\pm$0.015 & 0.454$\pm$0.009 & 0.402 $\pm$0.039 \\ 
Ours (IRL: All - Exec.)$^\diamond$ & 0.427 $\pm$0.019 & \textbf{0.683} $\pm$0.014 &0.437  $\pm$0.019 & \textbf{1.000} $\pm$0.000 \\
Ours (IRL: All) & 0.428 $\pm$0.027 & 0.693 $\pm$0.029 & 0.437  $\pm$0.032 &0.410 $\pm$0.029 \\
Ours (IRL: All)$^\diamond$ & 0.424 $\pm$0.022 & 0.687 $\pm$0.023 & 0.432 $\pm$0.024 & \textbf{1.000} $\pm$0.000 \\
\midrule
\rowcolor{GreenRow}
\multicolumn{5}{c}{\textbf{221 training samples}}\\

MLE &  0.283 $\pm$0.007 & 0.837 $\pm$0.005 & 0.234  $\pm$0.012 & 0.282 $\pm$0.049 \\

RL (LCS) &  0.324 $\pm$ 0.016 & 0.796  $\pm$0.010 & 0.272 $\pm$0.031 & 0.286 $\pm$0.037 \\

RL (LCS))$^\diamond$ & 0.316 $\pm$0.012 & 0.799 $\pm$0.010 & 0.276  $\pm$0.018 & \textbf{1.000} $\pm$0.000 \\

RL (LCS + exec) & 0.319 $\pm$0.018 & 0.798 $\pm$0.014 & 0.276 $\pm$0.025 & 0.309 $\pm$0.038 \\

RL (LCS + exec))$^\diamond$ & 0.313 $\pm$0.016 & 0.801 $\pm$0.011 & 0.271 $\pm$0.019 & \textbf{1.000} $\pm$0.000 \\

RL (All - Exec.) &   0.322 $\pm$0.018 & 0.802 $\pm$0.014 & 0.290 $\pm$0.024 & 0.289 $\pm$0.054 \\
RL (All) &  0.328 $\pm$0.009 & 0.796 $\pm$0.009 & 0.299 $\pm$0.016 & 0.273 $\pm$0.025 \\
Ours (IRL: All - Exec.) &  \textbf{0.342$^{*}$} $\pm$0.013 & \textbf{0.780} $\pm$0.014 & \textbf{0.312$^{*}$} $\pm$0.018 & 0.299 $\pm$0.043 \\
Ours (IRL: All - Exec.)$^\diamond$ &  0.336 $\pm$0.010 & \textbf{0.780} $\pm$0.010 & 0.305 $\pm$0.012 & \textbf{1.000} $\pm$0.000 \\

Ours (IRL: All) &   0.334 $\pm$0.018 & 0.791 $\pm$0.019 & 0.301 $\pm$0.024 & 0.286 $\pm$0.032 \\

Ours (IRL: All)$^\diamond$ &  0.327 $\pm$0.018 & 0.790 $\pm$0.015 & 0.294  $\pm$0.024 & \textbf{1.000} $\pm$0.000 \\

\midrule
\rowcolor{GreenRow}
\multicolumn{5}{c}{\textbf{70 training samples}}\\

MLE & 0.190 $\pm$0.009 & 0.919 $\pm$0.005 & 0.121 $\pm$0.006 & 0.221 $\pm$0.066 \\

RL (LCS) & 0.231  $\pm$0.005 & 0.892 $\pm$	0.006 & 0.158 $\pm$	0.010 & 0.152 $\pm$	0.047\\

RL (LCS)$^\diamond$) & 0.221 $\pm$0.005 & 0.895 $\pm$0.007 & 0.145 $\pm$0.010 & \textbf{1.000} $\pm$0.000 \\

RL (LCS + exec) &  0.223 $\pm$	0.009	& 0.900 $\pm$0.008	& 0.146 $\pm$0.011 & 0.191 $\pm$	0.044\\

RL (LCS + exec))$^\diamond$ & 0.216 $\pm$0.008 & 0.899 $\pm$0.007 & 0.137 $\pm$0.011 & \textbf{1.000} $\pm$0.000 \\

RL (All - Exec.) &  0.226 $\pm$0.012 & 0.899 $\pm$0.007 & 0.156 $\pm$0.017 & 0.142 $\pm$0.035 \\

RL (All) & 0.224 $\pm$0.014 & 0.896 $\pm$0.011 & 0.155 $\pm$0.018 & 0.182 $\pm$0.033 \\

Ours (IRL: All - Exec.) &  \textbf{0.246$^{*}$} $\pm$.008	& 0.881 $\pm$0.009 & \textbf{0.180$^{*}$} $\pm$	0.009 & 0.152 $\pm$	0.039\\ 

Ours (IRL: All - Exec.)$^\diamond$ &  0.236 $\pm$0.008 & 0.883 $\pm$0.006 & 0.166 $\pm$0.004 & \textbf{1.000} $\pm$0.000 \\

Ours (IRL: All) & 0.241 $\pm$	0.019 & \textbf{0.745$^{*}$} $\pm$	0.297 & 0.176 $\pm$	0.024 & 0.148 $\pm$	0.027 \\

Ours (IRL: All)$^\diamond$ &  0.234 $\pm$0.020 & 0.880 $\pm$0.014 & 0.165 $\pm$0.021 & \textbf{1.000} $\pm$0.000 \\

\midrule

\rowcolor{GreenRow}
\multicolumn{5}{c}{\textbf{Semi-supervised learning}}\\

Sup. \#70 + Unsup. \#697 & 0.127 $\pm$0.023 & 0.968 $\pm$	0.011 & 0.063 $\pm$	0.026	&  0.103 $\pm$0.017 \\
\bottomrule
\end{tabular} 
\caption{Test set performance for different training data sizes. MLE, RL(LCS), RL(LCS + exec) models are adapted from~\newcite{puig2018virtualhome}. $\diamond$ denotes executability as hard constraint during inference. * denotes statistical significance ($p < 0.01$) 
of IRL vs RL baselines from \newcite{puig2018virtualhome} using the Wilcoxon signed-rank test. 
} 
\label{tab:results_test_697}
\end{table*}


Table~\ref{tab:results_test_697} shows a comparative evaluation of our IRL approach with baseline methods. As baselines, we use (1) Random command generation, (2) MLE approach from \newcite{puig2018virtualhome}, (3) RL-based fine-tuning approach from \newcite{puig2018virtualhome} using LCS and execution rewards and (4) RL using all reward components (and a variant excluding executability component) weighted uniformly. 
We explore two variants of IRL: using all reward components, and all except the execution reward component. 
The table shows results with two types of inference: (1) we perform beam search with beam-size of 3 and (2) we enforce executability as a hard constraint (rows in the table marked with $\diamond$) while doing beam search, i.e., at any step, non-executable programs are dropped from the beam.
We can make the following conclusions on the quantitative performance of IRL over baselines:
\begin{itemize} [noitemsep, topsep=0pt, leftmargin=*]
    \item \textbf{Significant improvement over prior work:}
    Table~\ref{tab:results_test_697} shows that IRL achieves relative improvement of 9\% on the mean LCS score of the test set against RL(LCS) \cite{puig2018virtualhome} when using 697 training samples. The gains over RL (LCS) are statistically significant (p < 0.01). The trend remains same even with reduced training set size.

\item \textbf{Better gains in low-data regime:}
IRL outperforms RL significantly when not using the execution reward for all training set sizes in mean LCS and recall-based metrics. On using all the reward components the performances of RL and IRL models are comparable when the training data size is largest (697) even though IRL is slightly better in mean LCS score. However, we find that IRL's improvement over RL gets increasingly larger as the training data size drops. In the low data regime (training data size of 70), we find that using IRL is more effective and leads to higher gains than RL.

\item \textbf{Improvement in other rewards:}
Table~\ref{tab:results_test_697} shows that IRL helps to get better recall from program in addition to better LCS score. The gains are increased as training data gets reduced.

\item \textbf{Multiple reward components leads to better programs:}
IRL (without execution reward component) improves significantly (p $<$ 0.01, Wilcoxon signed-rank test) on recall from program and mean LCS scores as compared to RL baseline (using only LCS). This is an interesting result, since the RL model optimizes directly for an LCS reward. This may suggest that the composite reward function in IRL might be leading to better optimization trajectories. In many scenarios, having an exact execution reward is not feasible (often due to lack of a robust simulator). We find that the execution reward signal does not contribute much, and other forms of supervision from the expert programs lead to better performance.

\item \textbf{Automatic weight learning through IRL is helpful:}
To judge usefulness of the weights of each reward component learned by IRL we implement a RL baseline, RL (All), with all the reward components weighted uniformly and compare it  with the IRL models.
We observe that doing IRL outperforms this baseline in mean LCS score irrespective of the training set size considered. When using all reward components except execution the relative gain achieved by IRL against RL in mean LCS score is 1.16\%, 6.21\% and 8.85\% when trained with 697, 221 and 70 training samples respectively. Thus we see the learning the weights of each reward component proves more fruitful when we reduce the number of training samples.


\item \textbf{Semi-supervised learning:} We perform an experiment to explore unsupervised reward components that don't depend on the ground truth program (last row in Table~\ref{tab:results_test_697}). We fine-tune the supervised model trained using 70 training samples by using only three unsupervised reward components - irrelevance, recall from description, and repetition. Fine-tuning is done on all 697 training samples. 
In this case, we note that irrelevance reward improves from -0.479 to -0.397 and recall from description reward improves from 0.104 to 0.111 
as compared to values from MLE model trained on 70 samples. However, the absence of a signal for enforcing sequential structure degenerates the predicted sequence of commands, and LCS score drops significantly.

This scenario also tests generalization as many commands in test will be unseen in the ground-truth programs corresponding to the small subset of training samples (72.9\% of the commands in the test set are unseen). However, to test true generalization, IRL-based models need to be tested in unseen scenarios (possibly in new environments, with new objects, unseen tasks and new compositions of individual commands). We do not explore this direction in this work.
\end{itemize}

\begin{table*}[!ht]
\centering
\small
\begin{tabular}{lcccc}
\toprule
                                 \textbf{\textsc{Model}}   &              
                  \textbf{\shortstack{ \textsc{Mean} \\ \textsc{LCS}} } &
                  \textbf{\textsc{ED} } &
                  \textbf{\shortstack{ \textsc{Recall} \\ \textsc{Prog.} } } &
                  \textbf{\textsc{Exec.} } 
\\
\midrule

RL (LCS) & 0.265 $\pm$0.006 & 0.869 $\pm$0.005 & 0.198 $\pm$0.008 & 0.112 $\pm$0.024 \\

Irrel. + Recall desc. & 0.081 $\pm$0.015 & 0.984 $\pm$0.005 & 0.033 $\pm$0.008 & 0.210 $\pm$0.032 \\

Irrel. + Recall desc. + Rep. &  0.147 $\pm$0.034 & 0.947 $\pm$0.014 &  0.074 $\pm$0.030 & \textbf{0.301} $\pm$0.301   \\

LCS + Irrel. &  0.267 $\pm$	0.005 & 0.868 $\pm$	0.010 & 0.192  $\pm$	0.007	& 0.131 $\pm$0.031\\

LCS + Recall desc. & 0.271 $\pm$0.010 & 0.862 $\pm$0.011 & 0.201 $\pm$0.016 & 0.139 $\pm$0.012 \\

LCS + Rep. & 0.271 $\pm$0.010 & 0.868 $\pm$0.008 & 0.202 $\pm$0.013 & 0.130 $\pm$0.041 \\

LCS + Recall from prog &  0.274 $\pm$0.009 & 0.864 $\pm$0.008 & 0.209 $\pm$0.013 & 0.095 $\pm$0.020 \\

All reward comp. - Exec &  0.273 $\pm$0.007 & 0.865 $\pm$0.003 & 0.213 $\pm$0.010 & 0.105 $\pm$0.033 \\

All reward comp. &  \textbf{0.282} $\pm$0.003 & \textbf{0.857} $\pm$0.006 & \textbf{0.217} $\pm$0.005 & 0.114 $\pm$0.018 \\
\bottomrule
\end{tabular}
\caption{Ablation results on validation set when trained on training set of size 70}
\label{tab:ablation}
\end{table*}

\noindent \textbf{Weights Analysis}: We analyze relative weights learned by IRL and find that it assigns highest weights to the LCS (normalized weight value of 0.50) and recall from program reward components (normalized weight value of 0.35). The next largest weights are for the repetition and execution components (normalized weight values of 0.14 and 0.06, respectively). We see that the irrelevance penalty gets assigned a small negative weight, thus, allowing a small number of unrelated objects to show up in the program as long as the LCS and recall scores are not affected. 
Qualitatively, in many instances, IRL fine-tuning enables identification of the correct action verb when choosing the command as opposed to RL. 
High reward weights specifically to LCS and recall from program help to improve ordering of commands and prevent unrelated objects from appearing in the programs.
\\
\noindent \textbf{Ablation Study}: 
Table \ref{tab:ablation} shows performance for ablations grouping various reward components during IRL.  When only irrelevance, recall from description, and repetition are used the training is often unstable but can have higher executability owing to generation of empty programs which are trivially executable. We observe that without LCS as a reward component, program generation gradually degenerates.
Hence for other ablation experiments, we keep LCS and couple it with other rewards. IRL using LCS coupled with any other reward component shows improvements in the mean LCS score w.r.t RL (LCS) baseline. 
LCS coupled with either of the recall-based rewards performs better than other reward groups. Using all reward components we get the best scores on validation set except on executability metric.

\subsection{Human Evaluation}
We perform human evaluation using Amazon Mechanical Turk to explore qualitative differences between programs generated by our approach (IRL) and baseline RL approach from \newcite{puig2018virtualhome}. Turkers are given a task description followed by two program samples, one from each method (order is randomized). The turkers are asked to rate both the programs w.r.t a few criteria on a scale of 1 to 5 (where 5 denotes highest) and also choose a preferred program. The criteria for rating are:
\begin{itemize}[noitemsep,topsep=0pt,leftmargin=*]
    \item \textbf{Relatedness to description:} 
    relatedness of objects in the program w.r.t. the task description.
    \item \textbf{Human-likeness:} 
    how closely a generated program mimics human behavior.
    \item \textbf{Task completion:} 
    how much of the task is accomplished by the program.
\end{itemize}
\begin{table}[!ht]
    \centering
    \small
    \begin{tabular}{|c|c|c|}
        \hline
         \textbf{\textsc{Criterion}} & \textbf{RL} & \textbf{Ours (IRL)} \\
         \hline
         Relatedness &  3.09 & \textbf{3.18} \\
         Human-likeness & 3.41 & \textbf{3.50} \\
         Task completion & 2.84 & \textbf{2.95}\\
         \hline
         Pref. count (out of 250) & 106 &\textbf{ 144}\\
         \hline
    \end{tabular}
    \caption{Human evaluation results} 
    \label{tab:amt}
\end{table}
We rate batches of 
50 programs by 5 turkers. The programs are generated by RL(LCS) and IRL (All - Exec.) models fine-tuned on 697 training examples. 
Executability is not enforced during inference for both the models.
Table \ref{tab:amt} shows the human evaluation results. IRL programs are overall rated better for all three criteria. In general, IRL generates more relevant commands and prevents repetitions due to irrelevance and repetition reward components. Also, programs generated by the IRL model are preferred in 57.6\% of the cases (statistically significant at $p < 0.01$, Binomial test ). 

\section{Conclusion}
We explored an approach for incorporating diverse reward components in instruction following tasks. Such components can often be defined by a domain expert, and encode inductive biases about a problem. Since reward weights are learned, these models can be robust to spurious reward components. However, the issue of possibly adversarial reward components remains to be explored. While the approach requires access to expert examples, since we focus on scenarios involving RL-based fine-tuning, these are presumed to be already available. The approach can potentially generalize to other domains and applications, and can be fertile ground for directions of future research. 



\bibliography{anthology,custom}
\bibliographystyle{acl_natbib}

\appendix
\section{Appendix}
\subsection{Training Details}
\paragraph{Model parameters}
The task description is tokenized into words. We do not remove stop words or lemmatize words. Embeddings for words are obtained by using pretrained word2vec \cite{word2vec} vectors (300 dimensional). Next, these embeddings are passed into an encoder RNN made of LSTM cells. The LSTM network is unidirectional with hidden dimension of 100. The decoder RNN is also unidirectional with hidden dimension as 100. The last hidden state of the encoder RNN is used to initialise the hidden state of the decoder.
We train each model for 400 epochs using the Adam optimizer \cite{adam}. We choose the hyperparameters and best epoch for each model by obtaining results on the validation set using beam size of 3 and not enforcing executability. Since we adapt the model from \newcite{puig2018virtualhome} the size of network is still same with around 3M parameters.

\paragraph{Hyper-parameter tuning}
 We use five different random seeds for five trials of each experiment: 42, 101, 123, 2020 and 2021. Batch size and learning rate are manually tuned in the range \{64,128, 256\} and \{0.001, 0.0005, 0.0001\} respectively. Based on the results on validation splits we chose batch size as 256 and learning rate as 0.001 to report the results. We use a weight of 0.05 for entropy regularization during policy gradient after trying out weights in the range \{0.001, 0.005, 0.01, 0.05, 0.1, 0.5, 10 \}. We chose 0.05 as weight for entropy regularization on the basis of model performance on validation set which peaks for 0.05 when the model is trained on 70 samples. For Adam optimizer the value of coefficients used for computing running averages of gradient and its square are 0.9, 0.999 (default values as per Pytorch) respectively. We do not use weight decay during optimization. All the models are trained for 400 epochs. For MaxEnt IRL \cite{ziebart2008maximum}, we sample a subset of the training set for reward approximation stage (learning reward component weights). When we have a training set size of 70, we sample 5 expert and generated programs for this stage. Similarly we sample 50 for training set size of 221 and 150 for training set of 697 examples.  

\paragraph{Software and hardware specifications}
All the models are coded using Pytorch 1.4.0\footnote{\url{https://pytorch.org/}} \cite{pytorch} and related libraries like numpy \cite{numpy}, scipy \cite{scipy} etc. The graphical simulator for VirtualHome\cite{puig2018virtualhome} used in the paper is publicly available\footnote{\url{https://github.com/xavierpuigf/virtualhome}}.
We run all experiments on GeForce RTX 2080 GPU of size 12 GB. The system has 256 GB RAM and 40 CPU cores. The inference process is run in parallel on all the cores. For IRL fine-tuning on training set of size 70 it takes around 80 minutes for 400 epochs, which increases to 10 hours when fine-tuning on the training set of 697 samples. Doing RL fine-tuning also takes similar amount of time since the time required for just reward weight approximation is quite less.

\subsection{Validation Set Results}
We choose the best performing model given a training paradigm and also the set of parameter based on the performance in the validation set. Table \ref{tab:results_val_697} shows the results on the validation set. We use the same set of metrics being used for test set. We train and test five runs for each model. We perform significance testing using the Wilcoxon signed rank test. The values in the tables marked with \textbf{*} are statistically significant ($p < 0.05)$.

\begin{table*}[!ht]
\centering
\small
\begin{tabular}{lcccc}
\toprule
                                 \textbf{\textsc{Model}}   &
                  \textbf{\shortstack{ \textsc{Mean}\\\textsc{LCS}}} &
                  \textbf{\textsc{ED}} &
                  \textbf{\shortstack{\textsc{Recall} \\ \textsc{Prog.}}} &
                  \textbf{\textsc{Exec.}} 
                  \\ \midrule

Random &  0.069 $\pm$0.001 & 0.997 $\pm$0.002 &  0.007 $\pm$0.002 & 0.004 $\pm$0.005\\    
\midrule                                     
\rowcolor{GreenRow}
\multicolumn{5}{c}{\textbf{697 training samples}}\\
MLE & 0.412 $\pm$0.009 & 0.719 $\pm$0.010 & 0.407 $\pm$0.013 & 0.373 $\pm$0.039\\ 

RL (LCS)  & 0.453 $\pm$0.012 & 0.678 $\pm$0.018 &  0.472 $\pm$0.010 & 0.419 $\pm$0.060\\ 

RL (LCS)$^\diamond$ & 0.454 $\pm$0.011 & 0.668 $\pm$0.017 & 0.467 $\pm$0.016 & \textbf{1.000} $\pm$0.000 \\

RL (LCS + exec) & 0.449 $\pm$0.022 & 0.680 $\pm$0.027 &  0.456 $\pm$0.029 & 0.420 $\pm$0.039 \\

RL (LCS + exec)$^\diamond$ & 0.438 $\pm$0.016 & 0.679 $\pm$0.023 &  0.444 $\pm$0.018 & \textbf{1.000} $\pm$0.000 \\
RL (All - Exec.) & 0.453 $\pm$0.020 & 0.681 $\pm$0.020 & 0.491 $\pm$0.029 & 0.443 $\pm$0.013 \\
RL (All) & 0.453 $\pm$0.017 & 0.680 $\pm$0.020 & \textbf{0.492} $\pm$0.021 & 0.404 $\pm$0.046\\
Ours (IRL: All - Exec.) &  \textbf{0.460} $\pm$0.008 & 0.670 $\pm$0.009 & 0.481 $\pm$0.007 & 0.423 $\pm$0.04 \\ 

Ours (IRL: All - Exec.)$^\diamond$ & 0.457 $\pm$0.008 & \textbf{0.662} $\pm$0.009 & 0.458 $\pm$0.017 & \textbf{1.000} $\pm$0.000 \\

Ours (IRL: All) &  0.458 $\pm$0.022 & 0.672 $\pm$0.027 & 0.480 $\pm$0.031 & 0.431 $\pm$0.038 \\

Ours (IRL: All)$^\diamond$ & 0.459 $\pm$0.018 & \textbf{0.662} $\pm$0.024 & 0.475 $\pm$0.023 & \textbf{1.000} $\pm$0.000 \\
\midrule
\rowcolor{GreenRow}
\multicolumn{5}{c}{\textbf{221 training samples}}\\

MLE &  0.312 $\pm$0.010 & 0.818 $\pm$0.010 & 0.275 $\pm$0.013 & 0.252 $\pm$0.029 \\

RL (LCS) & 0.362 $\pm$0.006 & 0.771 $\pm$0.008 & 0.339 $\pm$0.007 & 0.308 $\pm$0.043 \\

RL (LCS))$^\diamond$ & 0.356 $\pm$0.010 & 0.779 $\pm$0.014 & 0.337 $\pm$0.014 & \textbf{1.000} $\pm$0.000 \\

RL (LCS + exec) & 0.358 $\pm$0.008 & 0.772 $\pm$0.012 & 0.329 $\pm$0.005 & 0.304 $\pm$0.042 \\

RL (LCS + exec))$^\diamond$ & 0.359 $\pm$0.008 & 0.762 $\pm$0.011 & 0.331 $\pm$0.007 & \textbf{1.000} $\pm$0.000 \\

RL (All - Exec.) & 0.353 $\pm$0.013 & 0.779 $\pm$0.010 & 0.340 $\pm$0.021 & 0.312 $\pm$0.035 \\

RL (All) & 0.359 $\pm$0.007	& 0.776 $\pm$0.008 & 0.345 $\pm$0.013 & 0.271	$\pm$0.039
\\ 

Ours (IRL: All - Exec.) & \textbf{0.376$^{*}$} $\pm$0.006 & 0.757 $\pm$0.007 & \textbf{0.362$^{*}$} $\pm$0.014 & 0.362 $\pm$0.034 \\

Ours (IRL: All - Exec.)$^\diamond$ & 0.368 $\pm$0.007 & \textbf{0.754} $\pm$0.006 & 0.354 $\pm$0.015 & \textbf{1.000} $\pm$0.000 \\

Ours (IRL: All) &  0.370 $\pm$0.006 & 0.767 $\pm$0.010 & 0.354  $\pm$0.011 & 0.279 $\pm$0.039 \\

Ours (IRL: All)$^\diamond$ & 0.370 $\pm$0.010 & 0.756 $\pm$0.010 & 0.357 $\pm$0.016 & \textbf{1.000} $\pm$0.000 \\
\midrule
\rowcolor{GreenRow}
\multicolumn{5}{c}{\textbf{70 training samples}}\\

MLE & 0.233 $\pm$0.015 & 0.866 $\pm$0.009 & 0.160 $\pm$0.011 & 0.232 $\pm$0.038 \\

RL (LCS) &  0.265 $\pm$0.006 & 0.869 $\pm$0.005 & 0.198 $\pm$0.008 & 0.112 $\pm$0.024 \\

RL (LCS)$^\diamond$) & 0.260 $\pm$0.010 & 0.865 $\pm$0.014 & 0.181 $\pm$0.017 & \textbf{1.000} $\pm$0.000 \\

RL (LCS + exec) &  0.261 $\pm$0.006 & 0.875 $\pm$0.009 & 0.185 $\pm$0.007 & 0.146 $\pm$0.023 \\

RL (LCS + exec))$^\diamond$ & 0.258 $\pm$0.006 & 0.868 $\pm$0.007 & 0.173 $\pm$0.005 & \textbf{1.000} $\pm$0.000 \\

RL (All - Exec.) & 0.267 $\pm$0.010 & 0.870 $\pm$0.011 & 0.198 $\pm$0.019 & 0.100 $\pm$0.050 \\

Rl (All) & 0.265 $\pm$0.006 & 0.871 $\pm$0.005 & 0.201 $\pm$0.008 & 0.113 $\pm$0.046\\

Ours (IRL: All - Exec.) & 0.273 $\pm$0.007 & 0.865 $\pm$0.003 & 0.213 $\pm$0.010 & 0.105 $\pm$0.033 \\

Ours (IRL: All - Exec.)$^\diamond$ & 0.276 $\pm$0.011 & 0.854 $\pm$0.017 & 0.213 $\pm$0.015 & \textbf{1.000} $\pm$0.000 \\

Ours (IRL: All) &  0.282 $\pm$0.003 & 0.857 $\pm$0.006 & \textbf{0.217$^{*}$} $\pm$0.005 & 0.114 $\pm$0.018 \\

Ours (IRL: All)$^\diamond$ & \textbf{0.283$^{*}$} $\pm$0.008 & \textbf{0.848$^{*}$} $\pm$0.013 & 0.207 $\pm$0.014 & \textbf{1.000} $\pm$0.000 \\

\midrule
\rowcolor{GreenRow}
\multicolumn{5}{c}{\textbf{Semi-supervised learning}}\\

Sup. \#70 + Unsup. \#697 & 0.145 $\pm$0.025 & 0.963 $\pm$	0.015 & 0.075 $\pm$	0.027 & 0.130 $\pm$	0.035\\
\bottomrule
\end{tabular}
\caption{Validation set performance for different training data sizes. MLE, RL(LCS), RL(LCS + exec) models are adapted from~\newcite{puig2018virtualhome}. $\diamond$ denotes executability as hard constraint during inference. * denotes statistical significance ($p < 0.05$) 
of IRL vs RL baselines from \newcite{puig2018virtualhome} using the Wilcoxon signed-rank test. 
} 
\label{tab:results_val_697}
\end{table*}

\subsection{Example of generated programs by IRL}
Table~\ref{tab:example_IRL} shows two sample generated programs from the IRL model trained with 697 samples using all rewards. The ``good" program has higher values of LCS and recall from program values.

\begin{table*}[!h]
    \centering
    \begin{tabular}{l}
         \toprule
         \textbf{Good Example}  \\
         \textbf{Task description:} \\
         I pick up groceries. I open fridge. 
         I put groceries in the fridge.\\
\textbf{Program:}\\
\text{[walk] $\langle$kitchen$\rangle$} \\
\text{[walk] $\langle$fridge$\rangle$} \\
\text{[find] $\langle$fridge$\rangle$}\\
\text{[open] $\langle$fridge$\rangle$}\\
\text{[find] $\langle$groceries$\rangle$}\\ 
\text{[grab] $\langle$groceries$\rangle$ }\\
\text{[putin] $\langle$groceries$\rangle$ $\langle$fridge$\rangle$ }\\
\text{[close] $\langle$fridge$\rangle$}\\
\\
\textbf{Rewards:} $\text{LCS} (0.50) , \text{Rec}_{\text{desc}} (1.00), \text{Rec}_{\text{prog}} (0.71), \text{Irrel.} (-0.12), \text{Exec.} (1)$ \\

\midrule

\textbf{Bad example:}\\
\textbf{Task description:}\\
Turn on laptop. Connect to internet and check for new email messages.\\
\textbf{Program:}\\
\text{ [walk] $\langle$home\_office$\rangle$}\\
\text{ [walk] $\langle$computer$\rangle$}\\
\text{[find] $\langle$computer$\rangle$}\\
\text{ [switchon] $\langle$computer$\rangle$}\\
\\
\textbf{Rewards:} $\text{LCS} (0.15) , \text{Rec}_{\text{desc}} (0.25), \text{Rec}_{\text{prog}} (0.08), \text{Irrel.} (0.0), \text{Exec.} (1)$\\
\bottomrule
    \end{tabular}
    \caption{Two sample generated programs with high and low values of LCS and Recall from program.}
    \label{tab:example_IRL}
\end{table*}




\end{document}